\documentclass[conference]{IEEEtran}
\IEEEoverridecommandlockouts
\usepackage{cite}
\usepackage{amsmath,amssymb,amsfonts}
\usepackage{graphicx}
\usepackage{textcomp}
\usepackage{todonotes}
\usepackage{algpseudocode}
\usepackage{algorithm}
\usepackage{hyperref}
\usepackage{bm}

\usepackage{xcolor}
\def\BibTeX{{\rm B\kern-.05em{\sc i\kern-.025em b}\kern-.08em
    T\kern-.1667em\lower.7ex\hbox{E}\kern-.125emX}}
\begin{document}

\title{Federated Learning for Anomaly Detection in Maritime Movement Data\\
\thanks{This work is mainly funded by the EU's Horizon Europe research and innovation program under Grant No. 101070279 MobiSpaces.}
}

\author{\IEEEauthorblockN{Anita Graser}
\IEEEauthorblockA{\textit{AIT Austrian Institute of Technology} \\
Vienna, Austria \\
anita.graser@ait.ac.at \\
ORCID: 0000-0001-5361-2885}
\and
\IEEEauthorblockN{Axel Weissenfeld}
\IEEEauthorblockA{\textit{AIT Austrian Institute of Technology} \\
Vienna, Austria \\
axel.weissenfeld@ait.ac.at}
\and
\IEEEauthorblockN{Clemens Heistracher}
\IEEEauthorblockA{\textit{craftworks GmbH} \\
Vienna, Austria \\
ORCID: 0009-0000-6864-1531}
\and
\IEEEauthorblockN{Melitta Dragaschnig}
\IEEEauthorblockA{\textit{AIT Austrian Institute of Technology} \\
Vienna, Austria \\
ORCID: 0000-0001-5100-2717}
\and
\IEEEauthorblockN{Peter Widhalm}
\IEEEauthorblockA{\textit{AIT Austrian Institute of Technology} \\
Vienna, Austria \\
ORCID: 0000-0002-5074-5356}
%\and
}

\maketitle

\begin{abstract} This paper introduces M³fed, a novel solution for federated learning of  movement anomaly detection models. This innovation has the potential to improve data privacy and reduce communication costs in machine learning for movement anomaly detection. We present the novel federated learning (FL) strategies employed to train M³fed, perform an example experiment with maritime AIS data, and evaluate the results with respect to communication costs and FL model quality by comparing classic centralized M³ and the new federated M³fed. 
\end{abstract}


\section{Introduction}

The deployment of machine learning approaches in practice often faces issues of data availability and communication bandwidth bottlenecks. Particularly in the mobility domain, data is often privacy sensitive and / or the communication network may be unreliable or rate limited. One approach to address these issues is Federated Learning (FL) since it can mitigate privacy risks and reduce communication costs compared to traditional centralized machine learning~\cite{Kairouz2021}. 

Federated learning is a machine learning setting where clients (such as, for example, mobile, edge, or Internet of Things devices) collaboratively train a model under the orchestration of a central server, while keeping the training data decentralized~\cite{Kairouz2021}. This means that the clients build their own local models and only share model updates with the central server. (This avoids the drawback of traditional approaches that require that all data is sent to the central server for model training.) The FL server then aggregates the model updates into the global model and transfers the global model back to the clients for further training.

FL applications include, for example, keyboard apps and vocal classifier in iOS, finance risk prediction for reinsurance, electronic health records mining, smart manufacturing~\cite{Kairouz2021}, predictive maintenance~\cite{Pruckovskaja2023b}, as well as weather forecasting and air pollution monitoring~\cite{Saylam2023}.

FL has not yet received significant attention in Spatial and Mobility Data Science~\cite{Graser2022, Graser2023}. Noteworthy recent contributions are: 
a model for location prediction in human mobility~\cite{li2020},
the detection of anomalous vehicle trajectories at intersections using federated learning~\cite{Koetsier2022} and 
a spatial–temporal model for freight traffic speed forecasting~\cite{Shen2024} and
a multimodal transport demand forecasting network with attentive federated learning (MAFL) ~\cite{li2023}.
To the best of our knowledge, however, there are no FL models for movement anomaly detection in the maritime domain. 

This work introduces the novel federated learning model M³fed. M³fed builds on the M³ massive movement model~\cite{Graser2020} which enables centralized learning of movement patterns for movement anomaly detection~\cite{Graser2018}. 
The next section provides a quick introduction into M³ followed by the introduction of the novel M³fed model. 
In the experiment section, we apply M³fed to anomaly detection in maritime movement data and compare results (detected anomalies and communication costs) with the centralized model.

\section{Method}
% \todo[inline, color=green!40]{Test}

This section provides a quick overview of the original centralized M³ model, followed by an introduction of the novel M³fed model.

\subsection{M³ (Centralized) Model}

M³ is a spatially explicit machine learning model. It divides geographic space into a regular grid and employs a prototype approach which models the multivariate distribution of the motion state vector in each grid cell as a multi-dimensional Gaussian Mixture distribution (including, but not limited to location, direction, and speed)~\cite{Graser2020}. A Gaussian Mixture Model (GMM) consists of multivariate Gaussian distributions called mixture components $C$. Each component $c$ has its own set of parameters $\theta_c = \{\bm{\mu}_c, \bm{S}_c\}$, where $\bm{\mu}_c$ is the mean value vector and $\bm{S}_c$ is the covariance matrix of the multivariate Gaussian.
Each component of the mixture also has an associated non-negative mixing weight $\pi_c$. These weights sum to one. The probability density function for the Gaussian mixture model is given by:

\begin{equation}
p(\bm{x})=\sum_{c=1}^{K}\pi_c N(\bm{x};\bm{\theta}_c).
\end{equation}

The model training process is described in detail in~\cite{Graser2020}. In short: the model is trained incrementally. For each new movement data record $\bm{x}$, first, we compute the spatially (or spatiotemporally) matching grid cell and then we update the Gaussian mixture distribution of the matching cell. This update looks for the most similar prototype vector in the set $C$ of Gaussian components of the cell using:

\begin{equation}
\bm{\mu}^*=\arg \min_{c \in C} \| \bm{x}-\bm{\mu}_c \|
\end{equation}

If the distance between the new movement data record and the most similar prototype is below the specified threshold, we update the prototype's mean vector $\bm{\mu}^*$, covariances $\bm{S}^*$, and number of data points. Otherwise, a new prototype is added. If the number of prototypes in a cell exceeds the specified limit, the most similar prototypes are merged. 

\subsection{Inference}

When the model is trained, it is ready to detect anomalies by comparing new, previously unseen movement data records to the model~\cite{Graser2018}. For how long or with how much input data the model has to be trained depends on the complexity of the movement patterns in the observed region. 

Movement data records are compared to nearby model prototypes.  If no  prototypes exist nearby, the record is considered a ``position anomaly'' since there is no past observation in the region. Otherwise, we select the most similar prototype to the movement data record and perform a chi-squared test to assess the goodness of fit of the new record to the modelled distribution. Records with a bad model fit are also reported as anomalies. 

Examining the deviation between record and model along each dimension of the motion state vector makes it possibly to give a reason for the anomaly, such as too low or too high speed (``speed anomaly'') or unusual course (``direction anomaly'').

\subsection{M³ Federated (M³fed)}

M³fed is a federated learning model. Our implementation is based on the Flower framework~\cite{beutel2020Flower}. We chose Flower since it has several advantages over other FL frameworks: it is open source and supports multiple programming languages and is ML framework-agnostic. Moreover, it enables federated learning on simulated as well as on distributed real world devices. However, other FL frameworks could be used instead, as long as they support the M³fed model described here.
%\todo[inline]{CH: I wouldn't call the FL strategy a component and adapted the text. - AG: done}

The FL framework encompasses clients, servers, and the federated learning strategy. FL clients train models locally, on their respective datasets, without sharing training data with a central server. The FL server orchestrates the training process across clients. It aggregates the model updates it receives from multiple clients, resulting in a global model. Finally, the federated learning strategy defines the protocol for communication and collaboration between clients and the server as well as the aggregation of the locally trained models. In our implementation, the communication between clients and server is based on gRPC. 

The training of the M³fed model consists of multiple steps as outlined in Algorithm~\ref{alg:m3fed}. Initially, an empty M³ model is created at the server, denoted as global model $\theta_0$. The federated training starts with initially transmitting the global model $\theta_{0}$ to the clients. Then, the clients carry out the local training using their own local records based on the empty M³ model resulting in $\theta_t^n$ (with $t$ being the index of the training round and $n$ defining the client). %The clients' training can be orchestrated by the server, too.

%The training of the M³fed model consists of multiple steps as outlined in Algorithm~\ref{alg:m3fed}. Initially, an empty M³ model is created at the server, denoted as global model $\theta_0$. A new round starts with transmitting the global model $\theta_{t-1}$ (with $t$ being the index of the round) to the clients. Then, the clients carry out the local training using their own local records %based on the global model 
%resulting in $\theta_t^n$ ($n$ defined the client). The clients' training can be orchestrated by the server, too.

%Afterwards, for each client, a ($\Delta \theta^n$) model is calculated by the difference of the newly trained local model and the previously received global model. These ($\Delta \theta^n$) models 
%\todo[inline]{CH: In my implementation, the global model is not part of the aggregation. Deleted global model in next sentence and line -1 in algorithm.}
The $\theta_t^n$ models are transmitted to the server, which aggregates the local models based on a previously implemented strategy resulting into a new global model $\theta_t$. During aggregation, the prototypes of all models are combined, which involves merging the most similar prototypes until the maximum number of prototypes is met
For details about the aggregation algorithm see~\cite{Graser2020}. Finally, the next training round starts. Since the transmission of models between clients and server is based on gRPC, the M³fed model is serialized in order to be adequate for transmission.

%\todo[inline]{CH:Maybe for global model $ t \in [0,T]$ would be more precises, since we mention the model at t = 0 earlier }

\begin{algorithm}
\caption{Training of M³fed}\label{alg:m3fed}
\begin{algorithmic}
\State Number of clients $N$ \newline
Number of rounds $T$ \newline
Data $D_t^n$ of client $n \in [1,N]$ and round $t \in [1,T]$ \newline
Client model $\theta_t^n$ at round $t \in [1,T]$ and client $n \in [1,N]$ \newline
Global model $\theta_t$ at round $t \in [1,T]$
\For{\texttt{t} $\gets$ \texttt{1} to \texttt{T} } 
  %\State $\theta_t^n \gets \theta_t$
  \For{\texttt{n} $\gets$ \texttt{1} to \texttt{N} } 
    %\State $\theta_{t+1}^n \gets$ train$(\theta_t^n, D_t^n)$ 
    \State $\theta_t^n \gets$ train$(\theta_0, D_t^n)$ \Comment{client side}
    %\State $\Delta \theta_{t+1}^n \gets$ $(\theta_t^n - \theta_t)$ \Comment{client side}
  \EndFor
  %\State $\theta_{t+1} \gets$ aggregation$(\theta_t, [\Delta \theta_{t+1}^n]_{n \in [1,N]})$ \Comment{server side}
  \State $\theta_{t} \gets$ aggregation$( [\theta_t^n]_{n \in [1,N]})$ \Comment{server side}
\EndFor
\end{algorithmic}
\end{algorithm}

%The strategy describes the aggregation algorithm of updates. The algorithm we have chosen will be discussed below in Sect. 4.7.2. The FL loop works as a central entity. It queries strategies, receives updates from connected clients, calculates global models and returns calculated models to clients via client manager. The client side only waits for instructions from the server and executes them. The client manager works with so called Flower messages. These messages are based on Remote Procedure Call ( RPC ) streams. RPC is efficient for low bandwidth transmission, because of the binary serialization format. So the client manager works as a RPC server for sending and receiving Flower messages [ BTM+20]. 

%Federated Learning M³fed

%\begin{itemize}
%\item Flower Framework 

%\item Initialization: empty prototype manager at server and clients:
%\item Fit function: add new prototypes from AIS
%\item Serialization

%\item Aggregation function: concat list of prototypes and load iteratively. 
%\item Communication protocol: GRPC
%\item Push new model to clients
%\end{itemize}

%Axel: Fragen
%\begin{itemize}
%\item LSTM-based model - predicting trajectory 4.2.3?
%\item Wie sollen die Daten auf train und test verteilt werden?
%\item stationary, mobile data - bedeutung?
%\end{itemize}

\section{Experimental set-up}

The experimental section of this paper uses open Automatic Identification System (AIS) data, which are available on the public infrastructure\footnote{\url{https://web.ais.dk/aisdata/}}. This data set consists of AIS records collected by the AIS receiver network of the Danish Maritime Authority and will be referred to as AISDK in this paper. 
For our experiments, we focus on the three ship types: \textit{cargo},  \textit{tanker} and  \textit{passenger} and, therefore, discard records associated with other ship types in the preprocessing step. In addition, incomplete records have been deleted.

\begin{figure}[tbp]
\centerline{\includegraphics[width=1\linewidth]{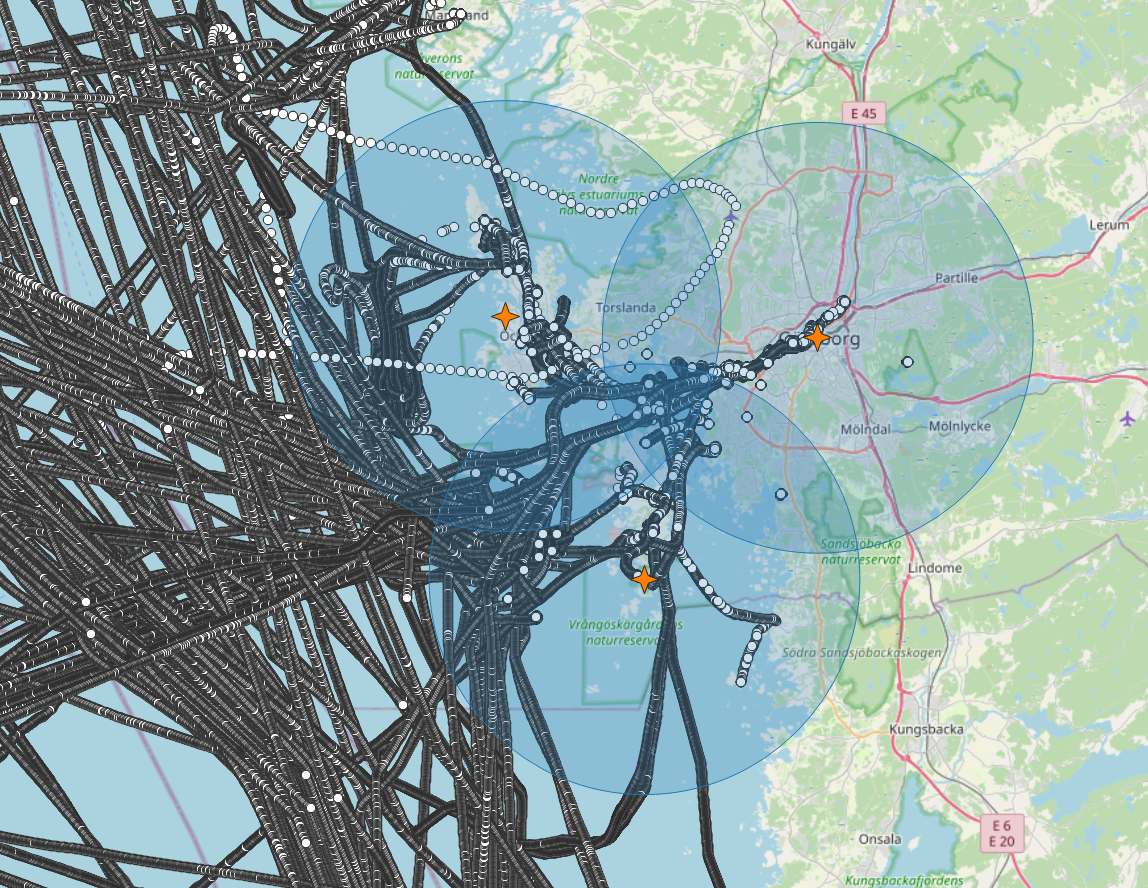}}
\caption{Locations of AISDK records (white points) and the three fictitious AIS antennas (orange stars) and their modelled overlapping coverage areas (blue circles) in the area around Gothenburg.}
\label{fig-antennas}
\end{figure}

We define three FL clients. Each client represents a fictitious AIS antenna that is receiving AIS records within its coverage area. Each FL client's coverage area encompasses a circular area defined by specific midpoints and a radius of about 13km. The clients are defined at (11.96524, 57.70730), (11.63979 57.71941) and (11.78460 57.57255). This simulation of AIS antennas is necessary since the AISDK dataset does not include antenna information. Figure~\ref{fig-antennas} shows the area of interest covered by our experiment.

We partition the data into training and test sets using a temporal split approach. The training data used in this study is based on one year of AIS data collected between July 2017 and June 2018. The total number of relevant AIS messages collected during this period is around 45 million %44,902,080 
messages. These records are assigned to the FL clients based on their location, as previously described. Client 1 has been allocated 16,880,947 records, client 2 has been allocated 25,701,110 records and client 3 has been allocated 23,447,490 records. The percentage distribution of training data among the clients is approximately 26\%, 39\%, and 35\%, respectively. Adding the records of all clients results in 66,029,547 records, which indicates that about 47\% of the records are assigned to more than one client since the coverage areas of the clients overlap. 

The training data is segmented based on dates, grouping all data from a single day into a training chunk for that particular training day. Consequently, each daily data chunk is allocated to a round in the federated learning approach. For instance, the training data between July 2017 and June 2018 encompasses 365 rounds.  

\begin{figure}[htbp]
\centerline{\includegraphics[width=1\linewidth]{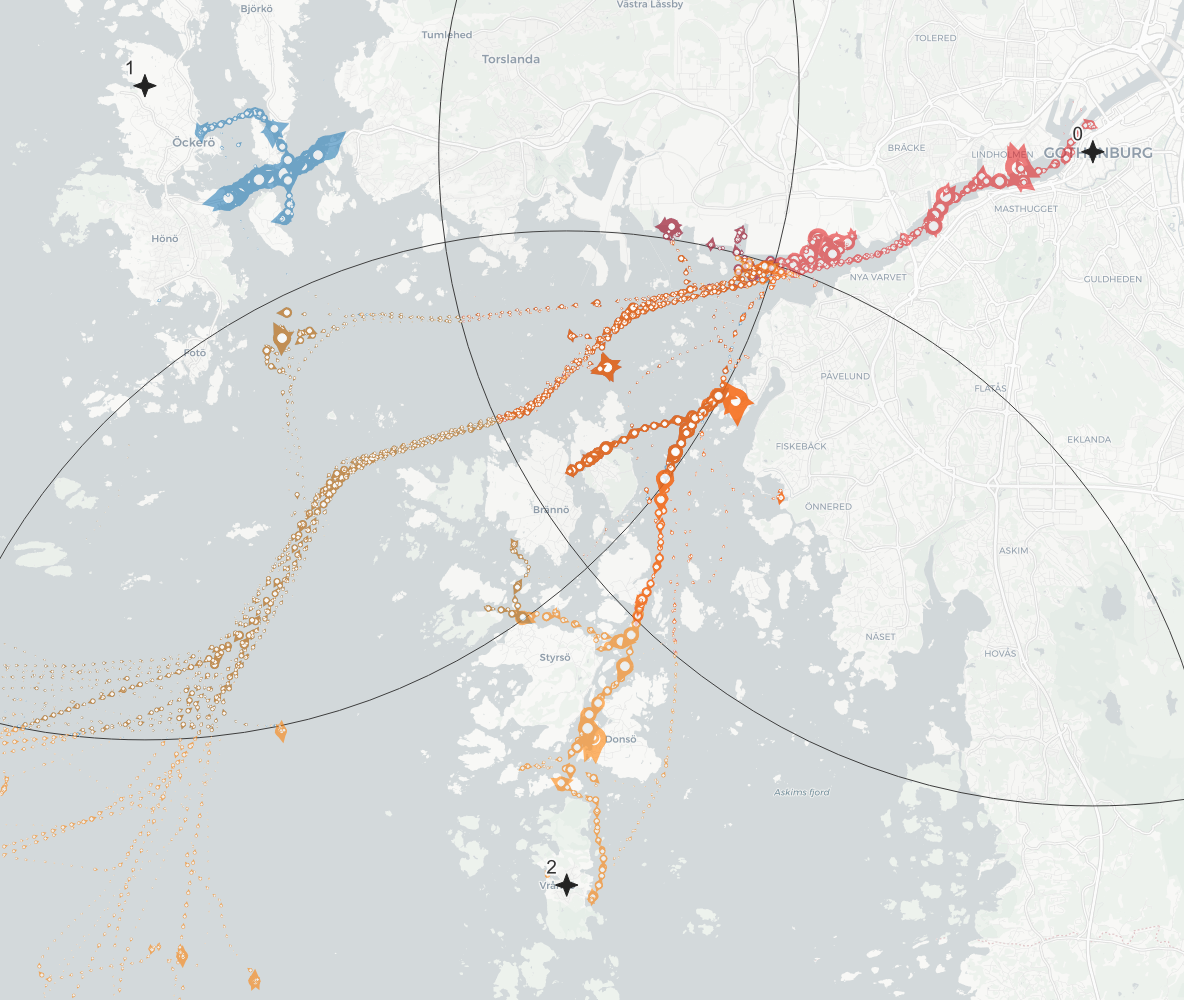}}
\caption{Prototypes learned by the three clients in round 1: red = client 0; blue = client 1; yellow = client 2. The size of the markers indicates the number of AIS records involved in training the respective prototype. }
\label{fig-round1}
\end{figure}

Figure~\ref{fig-round1} shows the prototypes trained on the clients in a single round. The geographic positions of the learned prototypes follow the vessel routes. The largest prototypes are found along popular routes and in ports. Where the coverage areas of the clients overlap, the learned prototypes overlap as well, leading to mixed colours (yellow, red, and blue) due to overlapping semi-transparent prototype markers in Figure~\ref{fig-round1}.

\begin{figure}[htbp]
\centerline{\includegraphics[width=1\linewidth]{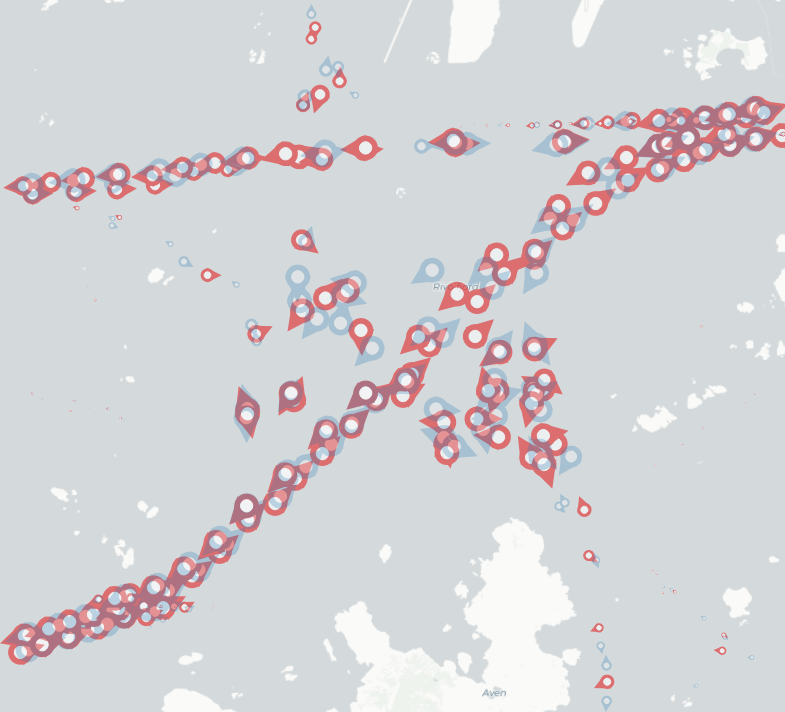}}
\caption{M³ (red) and M³fed (blue) prototypes in comparison. The size of the markers indicates the number of AIS records involved in training the respective prototype. }
\label{fig-comparison}
\end{figure}

Figure~\ref{fig-comparison} shows a comparison of the final trained M³ (red) and M³fed (blue) model prototypes in the central region where all three clients overlap. Areas where red and blue prototype markers overlap (resulting in purple markers) indicate high agreement between M³ and M³fed. Other areas contain clearly distinct red and blue prototype markers indicating that the model learned the movement patterns differently. However, to what degree these differences influence the anomaly detection, has to be determined in the following test / evaluation step.

The test data used in this study is a subset of the AIS data collected in July 2018. Out of the total number of AIS messages collected in July 2018, we removed incomplete records and excluded ship types that are not classified as cargo, tanker or passenger. This resulted in 4,540,004 records for testing our model. 
The thresholds for the anomaly detection are set to: position (pos) = 0.01, speed (sog) = 0.01, direction (cog) = 0.01.

%DATASETS
%\begin{itemize}
%\item Data Set: open-source AIS data from https://web.ais.dk/aisdata/
%\item temporal split between training and test data
%\item (only) stationary data
%\begin{itemize}
%\item training data: July 2017 - June 2018; num records: 66029547
%\item training data: total memory usage 29919276112bytes $\approx$29.92 GB 
%\item test data: July 2018; num records: 6911177 
%\end{itemize}
%\end{itemize}

%TRAINING
%\begin{itemize}
%\item Training
%\begin{itemize}
%\item central model is trained with all training data
%\item fl model is trained with all training data, which is distributed to the three clients
%\end{itemize}
%\item Evaluation
%\begin{itemize}
%\item RESULTS: TP 468390 TN 3146249 FP 0 FN 0 (ongoing)
%\end{itemize}
%\end{itemize}

\section{Results and Discussion}

In this section, we evaluate the results with respect to anomaly detection performance and communication costs comparing the M³ model and the M³fed model. We chose this approach since, to the best of our knowledge, there are no labelled benchmark datasets for movement anomaly detection in AIS data.

\subsection{Anomaly Detection Results}

Since there is no ground truth for actual anomalies in the dataset, we utilize the M³ model as the baseline and the M³fed model is expected to identify the same or at least similar anomalies. The evaluation of performance involves measuring true positives (TP), true negatives (TN), false positives (FP), and false negatives (FN). Table~\ref{tab:assessment} presents the results based on the test data of July 2018 and the previously stated anomaly detection thresholds.
%following results: 
%TP 91,368 (2.0\%); 
%TN 4,354,885 (95.9\%); 
%FP 31,828 (0.7\%) and 
%FN 61,331 (1.4\%).

\begin{table}[htbp]
    \caption{Anomaly detection assessment}
    \begin{center}
    \begin{tabular}{|c|c|c|}
    \hline
         &  M³fed anomaly& M³fed negative  \\
         \hline
         M³ anomaly &  TP = 91,368 (2.0\%)& FN = 61,331 (1.4\%)  \\
         M³ negative &  FP = 31,828 (0.7\%)& TN = 4,354,885 (95.9\%)  \\
         \hline
    \end{tabular}
    \label{tab:assessment}
    \end{center}
\end{table}

%\begin{table}[htbp]
%    \caption{Anomaly detection assessment}
%    \begin{center}
%    \begin{tabular}{|c|c|c|c|}
%    \hline
%         &  M³fed anomaly& M³fed negative & \\
%         \hline
%         M³ anomaly &  91,368 &  61,331  &   152,699 \\
%          &   TP = 2.0\% &  FN = 1.4\% & 3.4\% \\
%          \hline
%         M³ negative & 31,828  & 4,354,885   &  4,386,713  \\
%          &  FP = 0.7\%&  TN = 95.9\% & 96.6\% \\
%         \hline
%          & 123,196  & 4,416,216  & \\
%          & 2.7\% &  97.3\% & \\
%         \hline
%    \end{tabular}
%    \label{tab:assessment}
%    \end{center}
%\end{table}

\begin{figure}[htbp]
\centerline{\includegraphics[width=1\linewidth]{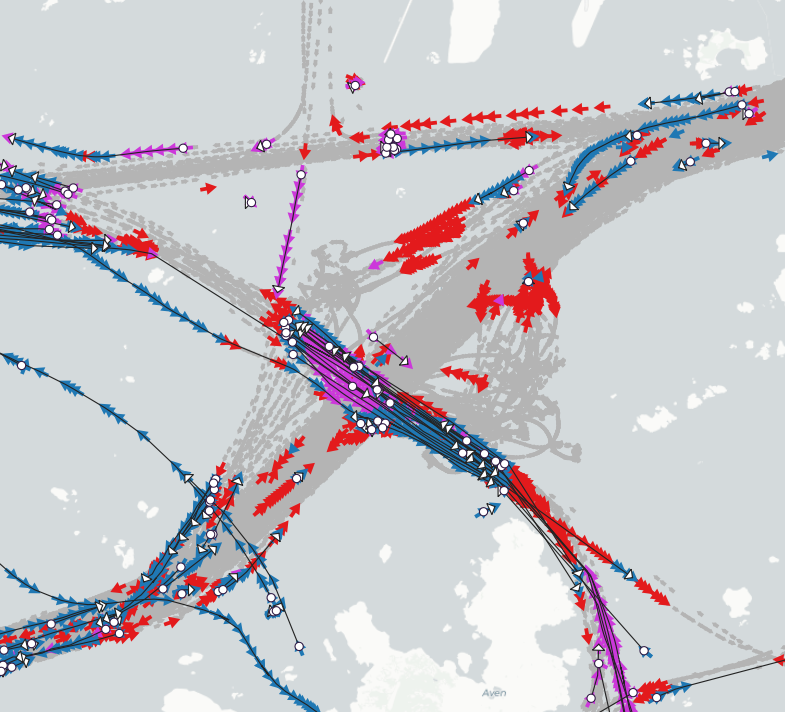}}
\caption{Anomaly assessment (comparison of anomalies from M³ and M³fed for ship type cargo): TP (blue), TN (gray), FP (pink), FN (red).}
\label{fig-anomalies-assess}
\end{figure}

Figure~\ref{fig-anomalies-assess} shows the assessed anomalies at a complex intersection of vessel routes in the center of the our area of interest where all three FL clients overlap. We can see where M³ and M³fed agree (blue and gray arrows) and where they disagree (red and pink arrows). 

\begin{figure}[htbp]
\centerline{\includegraphics[width=1\linewidth]{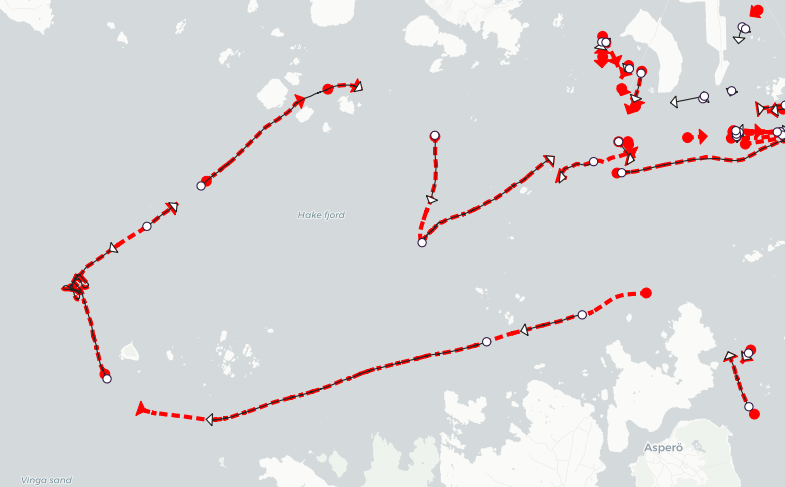}}
\caption{Anomaly events of vessel 265501910: M³fed events (black) on top of M³ events (red) illustrating the m:n relationships of anomaly events and the overlaps between events.}
\label{fig-anomaly-events}
\end{figure}

\begin{figure}[tbp]
\centerline{\includegraphics[width=1\linewidth]{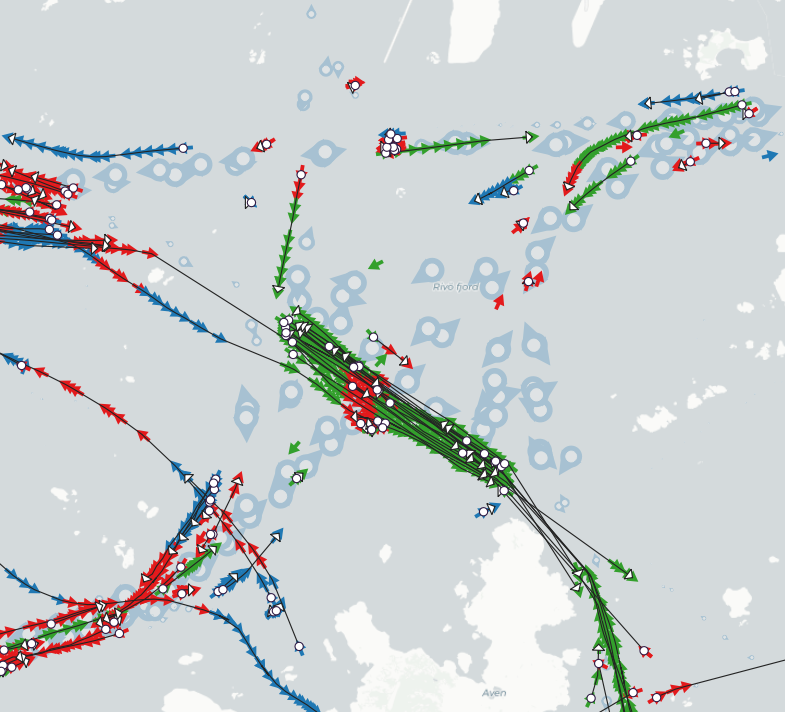}}
\caption{Anomaly types detected by M³fed  (for ship type cargo): position anomalies (blue), speed anomalies (green), and direction anomalies (red) overlaid on the M³fed prototypes.}
\label{fig-anomalies}
\end{figure}

\begin{figure}[htbp]
\centerline{\includegraphics[width=1\linewidth]{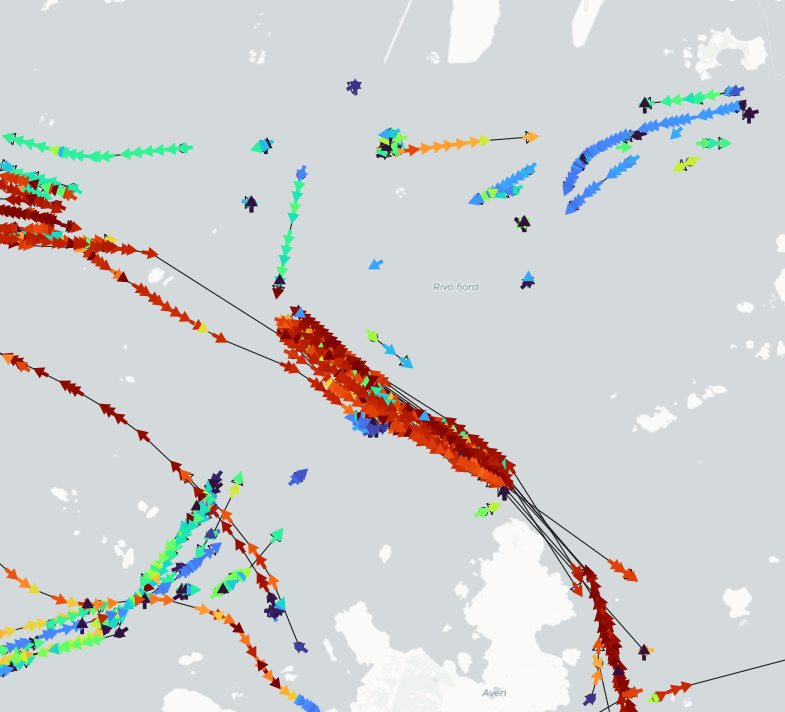}}
\caption{Anomaly speeds (for ship type cargo): colored by speed along blue-yellow-red (``Turbo'') gradient from 0 to 50km/h.}
\label{fig-anomalies-speed}
\end{figure}

An important fact to note in this evaluation is that AIS records are not independent. Instead, the records sent by a vessel are strongly spatiotemporally correlated. Additionally, from a user or application perspective, single anomalous records are not necessarily significant. Instead, a meaningful \textit{anomaly event} encompasses all related unusual AIS records. Therefore, for the purpose of this experiment, we define an anomaly event as all anomalous AIS records from the same vessel that can be connected with a maximum time gap of 1 minute between consecutive records (computed using the MovingPandas library~\cite{movingpandas}).  To illustrate this point, Figures~\ref{fig-anomalies-assess}-\ref{fig-anomalies-speed} show black lines that visualize the anomaly events by connecting consecutive anomaly records.

The \textit{durations of the anomaly events} exhibit a long-tailed distribution with many short anomaly events and few long event. Of the 12,156 anomaly events detected by M³fed, 6,966 (57\%) are shorter than 60 seconds and 161 (1\%) are longer than 10 minutes. 
By comparison, of the 15,933 anomaly events detected by M³, 10,009 (63\%) are shorter than 60 seconds and 160 (1\%) are longer than 10 minutes. 
The direct comparison of anomaly events is, however, complicated by the m:n relationships of M³ and M³fed-derived events, as shown in Figure~\ref{fig-anomaly-events}.

\begin{table}[htbp]
    \caption{Detected anomalies for July 2018 by type and model}
    \begin{center}
    \begin{tabular}{|l|c|c|}
    \hline
         & \textbf{M³} & \textbf{M³fed}\\
         \hline
 Position anomalies  &    35,228 &    29,011 \\
 Speed anomalies     &    45,056 &    37,862 \\
 Direction anomalies &    72,415 &    56,323 \\
 \hline
 No anomaly          & 4,312,088 & 4,416,216 \\
 \hline
 Total number of records & 4,539,412 & 4,539,412 \\
 \hline
    \end{tabular}
    \label{tab:anomalies}
    \end{center}
\end{table}

The detected anomalies can be further categorized by their main \textit{anomaly type} into: position, speed, and direction anomalies, as presented in Table~\ref{tab:anomalies}. 
Figure~\ref{fig-anomalies} shows the anomalies by type detected by M³fed. Direction anomalies (in red) show up where vessels cross the main vessel route (as indicated by the large prototype markers in the background). Position anomalies (in blue) show up in parts of the map where there have been no cargo vessels observed in the past and therefore no prototypes have been trained. Speed anomalies (in green) show up in locations where the expected speed is violated. For example, in Figure~\ref{fig-anomalies-speed}, we can see the high speeds of the anomalies in the map center (in red).  

\subsection{Communication Costs}

In real-world applications, the exchange of large datasets between clients and server can be problematic due to data protection concerns or bandwidth limitations or similar challenges. Since federated learning is often mentioned as potential solution for reducing the need for sharing large datasets since only models need to be exchanged, we also assessed the transmitted model sizes in relation to the volume of training data collected and utilized by the clients. If one round of the federated learning equals one day, then the total size of the exchanged model training data is around 0.44GB. This data includes the transmission of the updated models of the three clients to the server. In case the the aggregated global model is also returned to the clients then the exchanged data increases to around 3.2GB. If the clients instead upload their data to a central server for training, then the amount of data needed to be transmitted accumulates to approximately 23.4GB. By applying FL the amount of data to be transmitted can be significantly reduced by approximately 98\% (86\% if the aggregated models are shared with clients). In case a training round of the federated learning is not assigned to one day but increased to two days, then the transmission data size is also reduced by approximately 50\%, so that only 0.22GB (1.6GB in case the aggregated models are shared with the clients) data would need to be transmitted between the clients and server. In this way the volume of transmitted data can be further reduced, which is especially important if bandwidth is limited.

%Evaluation:
%\begin{itemize}
%\item Communication costs:
%\begin{itemize}
%\item Base line: centralized model with all AIS data set
%\item Our approach: only prototypes are sent: 
%Basically, we need to track the amount of data the server receives
%Potential improvement: send only changes in one direction, but models need to stay synchronized
%\end{itemize}
%\item	Model quality:
%\begin{itemize}
%\item Precision and recall vs central model on the same test data set 
%\end{itemize}
%\item	Montor system resources? https://Flower.dev/docs/framework/how-to-monitor-simulation.html 
%\end{itemize}

%Communication cost and quality over time

%Axel: Fragen
%\begin{itemize}
%\item evaluation method on client side not implemented yet - how is the model evaluated?
%\item ...
%\end{itemize}

\section{Conclusions and outlook}

This paper introduced a novel model for horizontal federated learning of movement anomalies called M³fed. We demonstrated M³fed's applicability using an example of AIS data and maritime movement anomalies and showed that M³fed provides comparable but not identical results to the centralized classic M³ model. 

A clear limitation of the experiments possible so far is the lack of ground truth data. Therefore, in future work, domain expert feedback will be necessary to determine which anomalies are considered relevant in practice. (An alternative approach, used in some papers, is to use simulated AIS with simulated anomalies. However, this approach comes with its own limitations since it remains unclear if these simulated AIS anomalies sufficiently reflect real-world vessel behavior.)

Since M³fed alleviates the need to share training data between clients and server, This reduces the communication bandwidth needs significantly, enabling anomaly model training in low and/or expensive bandwidth regions common in maritime use cases.

Further model tuning, particularly with regards to the maximum number of prototypes per cell parameter, will be performed to find the optimal trade-off between model size and detection results. Additional potential for compressing the model before transmission is to be investigated as well. 

M³fed is not limited to stationary federated learning clients. We are currently working on demonstrating M³fed for mobile clients, such as, for example, vessels moving on the open ocean which experience particularly severe limitations of their communication bandwidth.

%\begin{figure}[htbp]
%\centerline{\includegraphics{fig1.png}}
%\caption{Example of a figure caption.}
%\label{fig}
%\end{figure}

%\section*{Acknowledgment}
%\section*{References}

\vspace{12pt}

%\color{red}
%IEEE conference templates contain guidance text for composing and formatting conference papers. Please ensure that all template text is removed from your conference paper prior to submission to the conference. Failure to remove the template text from your paper may result in your paper not being published.

\end{document}